\documentclass{article} % For LaTeX2e
\usepackage{nips11submit_e,times}
\usepackage{epsfig}
\usepackage{graphicx}
\usepackage{multirow}
\usepackage{amsmath}

\title{Linearized Additive Classifiers}

\author{
Subhransu Maji\\
Department of Computer Science\\
University of California at Berkeley\\
\texttt{smaji@cs.berkeley.edu} 
}

\nipsfinalcopy % Uncomment for camera-ready version

\begin{document}
\maketitle

\begin{abstract}
We revisit the additive model learning literature and adapt a \emph{penalized spline} formulation due to Eilers and Marx~\cite{eilers2002generalized}, to train additive classifiers efficiently. We also propose two new embeddings based two classes of \emph{orthogonal basis with orthogonal derivatives}, which can also be used to efficiently learn additive classifiers. This paper follows the popular theme in the current literature where kernel SVMs are learned much more efficiently using a approximate embedding and linear machine. In this paper we show that spline basis are especially well suited for learning additive models because of their sparsity structure and the ease of computing the embedding which enables one to train these models in an online manner, without incurring the memory overhead of precomputing the storing the embeddings. We show interesting connections between B-Spline basis and histogram intersection kernel and show that for a particular choice of regularization and degree of the B-Splines, our proposed learning algorithm closely approximates the histogram intersection kernel SVM. This enables one to learn additive models with \emph{almost no memory overhead} compared to fast a linear solver, such as LIBLINEAR, while being only $5-6\times$ slower on average. On two large scale image classification datasets, \texttt{MNIST} and Daimler Chrysler pedestrians, the proposed additive classifiers are as accurate as the kernel SVM, while being two orders of magnitude faster to train. 
\end{abstract}

\section{Introduction}
Non parametric models for classification have become attractive since the introduction of kernel methods like the Support Vector Machines (SVMs)~\cite{boser1992training}. The complexity of the learned models scale with the data, which gives them desirable asymptotic properties. However from an estimation point of view, parametric models can offer significant statistical and computational advantages. Recent years has seen a shift of focus from non-parametric models to semi-parametric for learning classifiers. This includes the work of Rahimi and Recht~\cite{rahimi2008random}, where they compute an approximate feature map $\phi$, for shift invariant kernels $K(|x-y|) \sim \phi(x)'\phi(y)$, and solve the kernel problem approximately using a linear problem. This line of work has become extremely attractive, with the advent of several algorithms for training linear classifiers efficiently (for e.g. \texttt{LIBLINEAR}~\cite{fan2008liblinear}, \texttt{PEGASOS}~\cite{shalev2007pegasos}), including online variants which have very low memory overhead. 

Additive models, i.e., functions that decompose over dimensions $\left( f(x) = \sum_i f_i(x_i) \right)$,  are a natural extension to linear models, and arise naturally in many settings. In particular if the kernel is additive, i.e. $K(x,y) = \sum_i K_i(x_i,y_i)$
then the learned SVM classifier is also additive. A large number of useful kernels used in computer vision are based on comparing distributions of low level features on images and are additive, for e.g., histogram intersection and $\chi^2$ kernel~\cite{maji2009max}. This one dimensional decomposition allows one to compute approximate feature maps independently for each dimension, leading to very compact feature maps, making estimation efficient. This line of work has been explored by Maji and Berg~\cite{maji2009max} where they construct approximate feature maps for the $\min$ kernel, and learn piecewise linear functions in each dimension. For $\gamma$-homogenous additive kernels, Vedaldi and Zisserman~\cite{vedaldi2010efficient} propose to use the closed form features of Hein and Bousquet~\cite{hein2005hilbertian} to construct approximate feature maps. 

Smoothing splines~\cite{wahba1990spline} are another way of estimating additive models, and are well known in the statistical community. Ever since Generalized Additive Models (GAMs) were introduced by Hastie and Tibshirani~\cite{hastie1990generalized}, many practical approaches to training such models for regression have emerged, for example the P-Spline formulation of Eilers and Marx~\cite{eilers2002generalized}. However these algorithms do not scale to extremely large datasets and high-dimensional features typical of image or text classification datasets. 

In this work we show that the spline framework can used to derive embeddings to train additive classifiers efficiently as well. We propose two families of embeddings which have the property that the underlying additive classifier can be learned directly by estimating a linear classifier in the embedded space. The first family of embeddings are based on the Penalized Spline (``P-Spline") formulation of additive models (Eilers and Marx~\cite{eilers2002generalized}) where the function in each dimension is represented using a uniformly spaced spline basis and the regularization penalizes the difference between adjacent spline coefficients. The second class of embeddings are based on a generalized Fourier expansion of the function in each dimension. 

This work ties the literature of additive model regression and linear SVMs to develop algorithms to train additive models in the classification setting. We discuss how our additive embeddings are related to additive kernels in Section~\ref{sec:add_rkhs}. In particular our representations include those of~\cite{maji2009max} as a special case arising from a particular choice of B-Spline basis and regularization. An advantage of our representations is that it allows explicit control of the smoothness of the functions and the choice of basis functions, which may be desirable in certain situations. Moreover the sparsity of some of our representations lead to efficient training algorithms for smooth fits of functions. We summarize the previous work in the next section. 
\section{Previous Work}

The history of learning additive models goes back to~\cite{hastie1990generalized}, who proposed the ``backfitting algorithm" to estimate additive models. Since then many practical approaches have emerged, the most prominent of which is the Penalized Spline formulation (``P-Spline") proposed by~\cite{eilers2002generalized}, which consists of modeling the one dimensional functions using a large number of uniformly spaced B-Spline basis. Smoothness is ensured by penalizing the differences between adjacent spline coefficients. We describe the formulation in detail in Section~\ref{sec:p-spline}. A key advantage of this formulation was the whole problem could be solved using a linear system. 

Given data $(x^k,y^k), k = 1,\ldots,m$ with $x^k \in R^D$ and $y^k \in \{-1,+1\}$, discriminative training of functions often involve an optimization such as: 
\begin{equation}
\min_{f \in F} \sum_k l\left(y^k, f(x^k)\right) +\lambda R(f)
\end{equation}
where, $l$ is a loss function and $R(f)$ is a regularization term. In the classification setting a commonly used loss function $l$ is the hinge loss function:
\begin{equation}
l\left(y^k,f(x^k)\right) = \max\left (0, 1 - y^k f(x^k)\right)
\end{equation}

For various kernel SVMs the regularization penalizes the norm of the function in the implicit Reproducing Kernel Hilbert Space (RKHS), of the kernel. Approximating the RKHS of these additive kernels provides way of training additive kernel SVM classifiers efficiently. For shift invariant kernels, Rahimi and Recht~\cite{rahimi2008random} derive features based on Boshner's theorem. Vedaldi and Zisserman~\cite{vedaldi2010efficient} propose to use the closed form features of Hein and Bousquet~\cite{hein2005hilbertian} to train additive kernel SVMs efficiently for many commonly used additive kernels which are $\gamma$-homogenous. For the $\min$ kernel, Maji and Berg~\cite{maji2009max} propose an approximation and an efficient learning algorithm, and our work is closely related to this. 

In the additive modeling setting, a typical regularization is to penalize the norm the $d$th order derivatives of the function in each dimension, $i.e.$, $R(f) = \sum_i  \int_{-\infty}^{\infty}f_i^{d}(t)^2$. Our features are based on encodings that enable efficient evaluation and computation of this regularization. For further discussion we assume that the features $x^k$ are one dimensional. Once the embeddings are derived for the one dimensional case, we note that the overall embedding is concatenation of the embeddings in each dimension as the classifiers are additive.

%We are interested in additive classifiers where the overall function decomposes over dimensions. Let $f(x) = \sum_i f_i (x_i)$ be an additive function. Then given a labelled dataset  $(x^k,y^k), k = 1,\ldots,m$ with $x^k \in R^D$ and $y^k \in \{-1,+1\}$, the standard spline formulation can be written as : 
%\begin{equation}
%\min_{f \in F} \frac{1}{n}\sum_{i=1}^{n}l\left(y^k,f(x^k)\right) +\lambda \sum_{i=1}^D \int_{-\infty}^{\infty}f_i^{d}(t)^2 w_i(t)dt
%\end{equation}
%Here $f_i^d$ denotes the $d$th derivative of the function $f_i$, which is typically set to $d=1,2$ and $w_i(t):(-\infty,\infty) \rightarrow [0,\infty)$ is a weight function. Examples of weight functions are:
%\begin{eqnarray*}
%w_i(t) &=& 1,\mbox{ if $t \in [a_i,b_i]$}, w(t) = 0, \mbox{ otherwise}\\
% w_i(t) &=& e^{-t^2/2}
%\end{eqnarray*}

\section{Spline Embeddings}
\label{sec:p-spline}
Eilers and Marx~\cite{eilers2002generalized} proposed a practical modeling approach for GAMs. The idea is based on the representing the functions in each dimension using a relatively large number of uniformly spaced B-Spline basis. The smoothness of these functions is ensured by penalizing the first or second order differences between the adjacent spline coefficients. Let $\boldsymbol \phi(x^k) $ denote the vector with entries $\phi_i(x^k)$, the projection of $x^k$ on to the $i$th basis function. The P-Spline optimization problem for the classification setting with the hinge loss function consists of minimizing $c(\mathbf{w})$:
\begin{equation}
c(w) = \frac{\lambda}{2} \mathbf{w'}D_d'D_d\mathbf{w} + \frac{1}{n}\sum_k \max \left(0, 1 - y^k\left(\mathbf{w}'\boldsymbol \phi(x^k)\right)\right) 
\end{equation}
The matrix $D_d$ constructs the $d$th order differences of $\boldsymbol \alpha$:
\begin{equation}
D_d\boldsymbol \alpha = \Delta^d \boldsymbol \alpha 
\end{equation}
The first difference of $\boldsymbol \alpha$, $\Delta^1 \boldsymbol \alpha$ is a vector of elements $\alpha_i - \alpha_{i+1}$. Higher order difference matrices can be computed by repeating the differencing. For a $n$ dimensional basis, the difference matrix $D_1$ is a $(n-1)\times n$ matrix with $d_{i,i} = 1$, $d_{i,i+1}=-1$ and zero everywhere else. The matrices $D_1$ and $D_2$ are as follows: 
\[ D_1  = \left( \begin{array}{rrrrr}
1 & -1 &  & &  \\
 & 1 & -1 & &  \\
&   & \ldots  &   &   \\
&  &  & 1 & -1    \\
\end{array} 
\right) ;
D_2  = \left( \begin{array}{rrrrrr}
1 & -2 &1  & &  &\\
 & 1 & -2 & 1&  &\\
&   & \ldots  &   &  &  \\
&   &  & 1 & -2 & 1  \\
\end{array} 
\right)                  
\] 
\emph{To enable a reduction to the linear case we propose a slightly different difference matrix $D_1$. We let $D_1$ be a $n\times n$ matrix with $d_{i,i} = 1, d_{i,i-1} = -1$}. This is same as the first order difference matrix proposed by Eilers and Marx, with one more row added on top. The resulting difference matrices $D_1$ and $D_2 = D_1^2$ are both $n\times n$  matrices:
\[ D_1  = \left( \begin{array}{rrrrr}
1 &  &  & &  \\
-1 & 1 &  & &  \\
&  -1& 1 & &   \\
&   & \ldots  &   &   \\
&  &  -1& 1 &    \\
&  &  & -1 & 1    \\
\end{array} 
\right);
 D_2  = \left( \begin{array}{rrrrrr}
1 &  &  & &  &\\
-2 & 1 &  & &  &\\
&  1& -2 &1 & &  \\
&  & 1 &-2 &1 &  \\
&   & & \ldots  & &  \\
&  & & 1& -2 & 1    \\
\end{array} 
\right)
\] 
The first row in $D_1$ has the effect of penalizing the norm on the first coefficient of the spline basis, which plays the role of regularization in the linear setting (e.g. ridge regression, linear SVMs, etc). Alternatively one can think of this as an additional basis at left most point with its coefficient set to zero. \emph{The key advantage is that the matrix $D_1$ is invertible and has a particularly simple form which allows us to linearize the whole system}. We will also show in Section~\ref{sec:add_rkhs} that the derived embeddings also approximate the learning problem of kernel SVM classifier using the $\min$ kernel~($K_{\min}$) for a particular choice of spline basis.
\begin{equation}
K_{\min}(\mathbf{x},\mathbf{y}) = \sum_i \min(x_i,y_i)
\end{equation}

\begin{figure}
\begin{center}
\begin{tabular}{ccc}
  \includegraphics[width=0.3\linewidth]{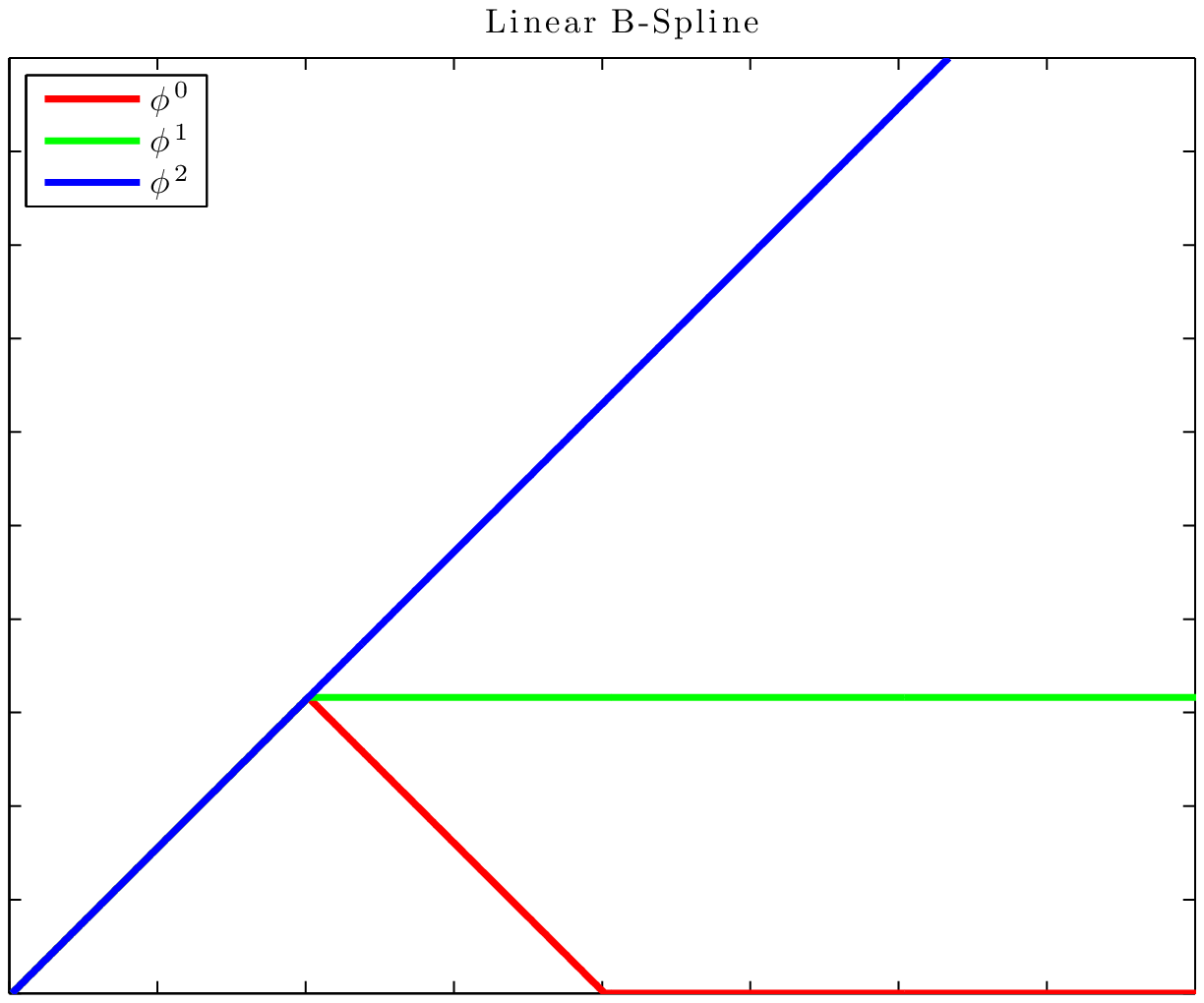}
&\includegraphics[width=0.3\linewidth]{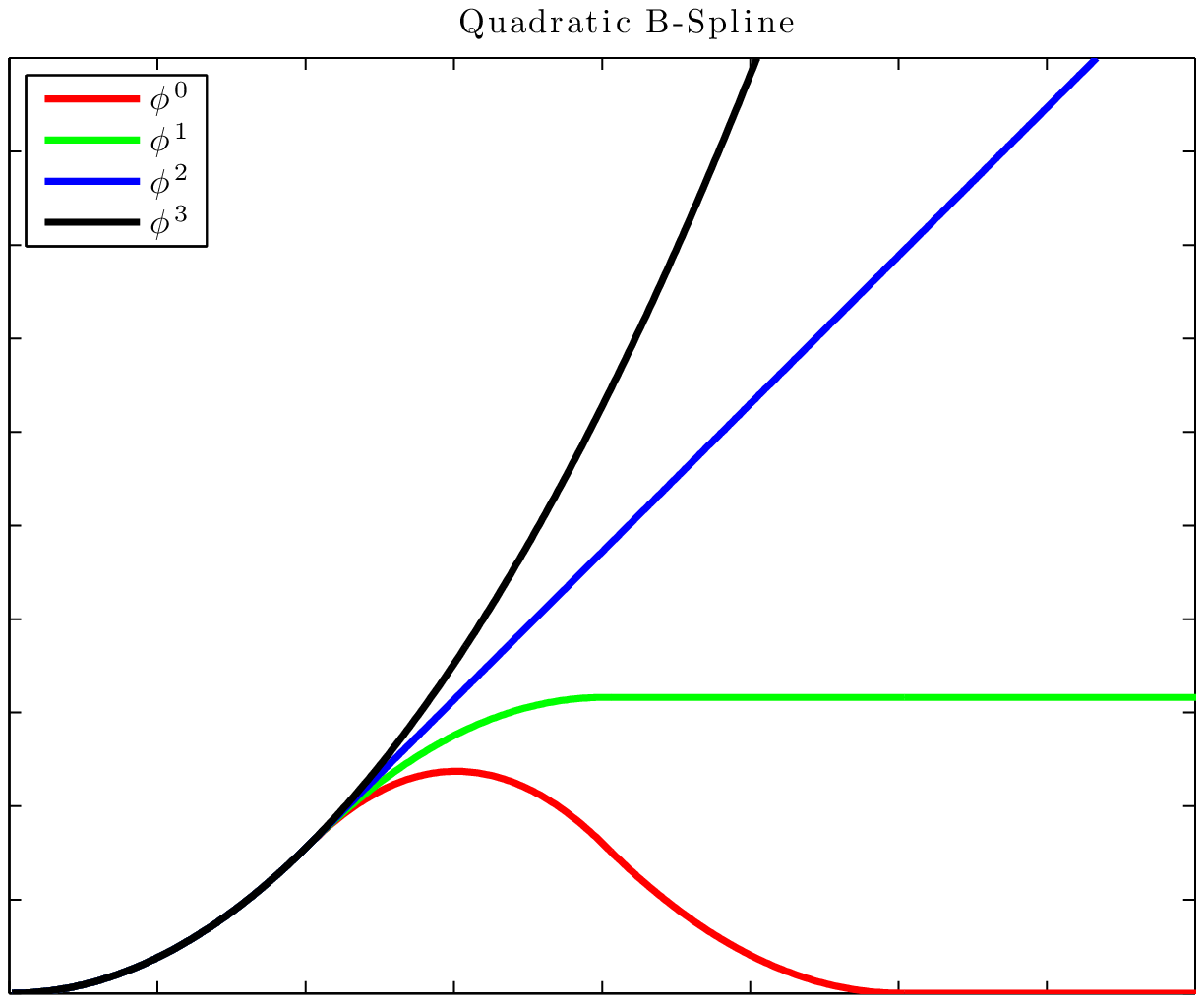}
&\includegraphics[width=0.3\linewidth]{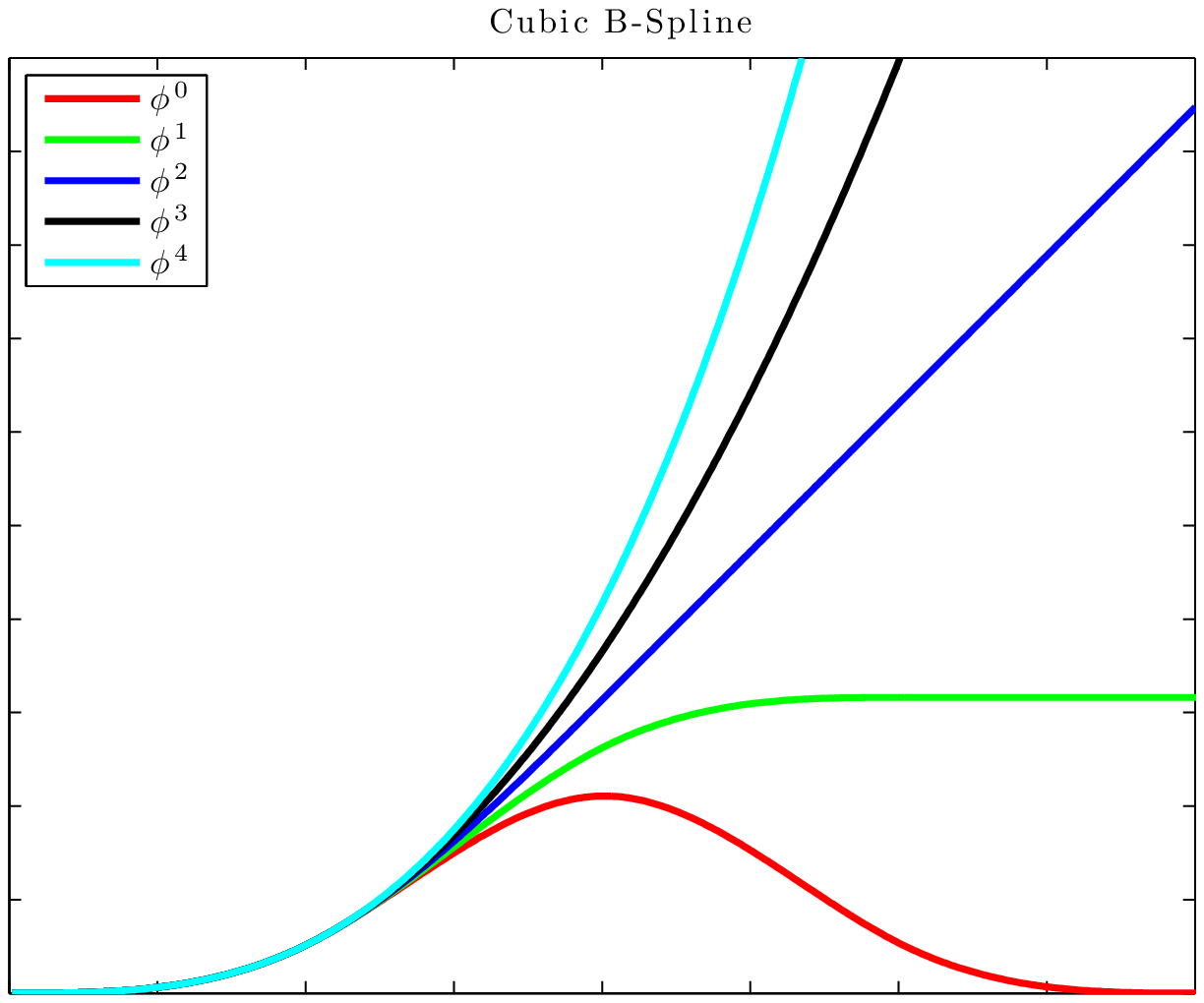}\\
\end{tabular}
\end{center}
\caption{\label{fig:bsplinebasis} Local basis functions for linear (left), quadratic (middle) and cubic (right) for various regularizations degrees $d$. In each figure $\boldsymbol \phi^d$ refers to the dense features $D_d^{'-1}\boldsymbol \phi$. When $d=0$, the function shown in the local basis of B-Splines. When $d=r+1$, where $r$ is degree of the B-Spline basis, then $\boldsymbol \phi^d$ are truncated polynomials basis,  $(x-\tau_i)_+^r$ (see Section~\ref{sec:add_rkhs}).}
\end{figure}

Given the choice of the regularization matrix $D_d$ which is invertible, one can linearize the whole system by re-parametrizing $\mathbf{w}$ by $D_d^{-1}\mathbf{w}$, which results in : 
\begin{equation}
c(w) = \frac{\lambda}{2} \mathbf{w}'\mathbf{w} + \frac{1}{n}\sum_k \max \left(0, 1 - y^k\left(\mathbf{w}'D_d^{'-1}\boldsymbol \phi(x^k)\right)\right) 
\end{equation}
Therefore the whole classifier is linear on the features $\boldsymbol \phi^d(x^k) = D_d^{'-1}\boldsymbol \phi(x^k)$, i.e. the optimization problem is equivalent to 
\begin{equation}
c(w) = \frac{\lambda}{2} \mathbf{w}'\mathbf{w} + \frac{1}{n}\sum_k \max \left(0, 1 - y^k\left(\mathbf{w}' \boldsymbol \phi^d(x^k)\right)\right) 
\end{equation}
The inverse matrices $D_1^{'-1}$ and $D_2^{'-1}$ are both upper triangular matrices. The matrix $D_1^{'-1}$ has entries $d_{i,j} = 1, j\geq i$ and $D_2^{'-1} = D_1^{'-2}$ has entries $d_{i,j} = j-i+1, j\geq i$ and look like:
\[ D_1^{'-1}  = \left( \begin{array}{cccccc}
1 & 1 &1 		& \ldots 	& 1 	& 1 \\
 & 1 & 1 		& \ldots 	& 1	& 1\\
   &  & 1 		& \ldots 	& 1 	& 1\\
   &    & \ldots  	& \ldots    &   \ldots	&  \\
   &    &           	&   		&  1	& 1\\
   &    &   		& 		&  	&1  \\
\end{array}\right);
 D_2^{'-1}  = \left( \begin{array}{cccccc}
1 & 2 &3 		& \ldots 	& n-1 	& n \\
 & 1 & 2 		& \ldots 	& n-2	& n-1\\
   &  & 1 		& \ldots 	& n-3 	& n-2\\
   &    & \ldots  	& \ldots    &   \ldots	&  \\
   &    &           	&   		&  1	& 2\\
   &    &   		& 		&  	&1  \\
\end{array} 
\right)\] 
We refer the readers to~\cite{eilers2005splines} for an excellent review of additive modeling using splines. Figure~\ref{fig:bsplinebasis} shows the $\boldsymbol \phi^d$ for various choices of the regularization degree $d=0,1,2$ and B Splines basis, linear, quadratic and cubic. 

\subsection{Generalized Fourier Embeddings}
Generalized Fourier expansion provides an alternate way of fitting additive models. Let $\psi_1(x), \psi_2(x),\ldots, \psi_n(x)$ be a orthogonal basis system in the interval $[a,b]$, wrto. a weight function $w(x)$,  i.e. we have $\int_a^b \psi_i(x) \psi_j(x) w(x)dx = 0, i\neq j$. Given a function $f(x)= \sum_i a_i \psi_i(x)$, the regularization can be written as: 
\begin{equation}
\int_{a}^bf^{d}(x)^2 w(x) dx = \int_{a}^b \left( \sum_i a_i \psi_i^d(x) \right)^2w(x)dx = \int_{a}^b \left( \sum_{i,j} a_i a_j \psi_i^d(x)\psi_j^d(x) \right)w(x)dx 
\end{equation}
\emph{Consider an orthogonal family of basis functions which are differentiable and whose derivatives are also orthogonal}. One can normalize the basis such that $\int_a^b \psi_i^d(x) \psi_j^d(x) w(x)dx = \delta_{ij}$. In this case the regularization has a simple form: 
\begin{equation}
\int_{a}^bf^{d}(x)^2 w(x) dx = \int_{a}^b \left( \sum_{i,j} a_i a_j \psi_i^d(x)\psi_j^d(x) \right)w(x)dx  = \sum_i a_i^2
\end{equation}
Thus the overall regularized additive classifier can be learned by learning a linear classifier in the embedded space $\psi(x)$. In practice one can approximate the scheme by using a small number of basis function. We propose two practical ones with closed form embeddings: 

\paragraph{Fourier basis.} The classic Fourier basis functions $\{1,\cos(\pi x),\sin(\pi x),\cos(2\pi x),\sin(2\pi x),\ldots\}$ are orthogonal in $[-1, 1]$, wrto. the weight function $w(x) = 1$. The derivatives are also in the same family (except the constant basis function), hence are also orthonormal. The normalized feature embeddings for $d=1,2$ are shown in Table~\ref{table:fourier_features}. 

\paragraph{Hermite basis.} Hermite polynomials also are an orthogonal basis system with orthogonal derivatives wrto. the weight function $e^{-x^2/2}$. Using the following identity: 
\begin{equation}
	\int_{-\infty}^{\infty} H_m(x) H_n(x) e^{-x^2/2} dx = \sqrt{2\pi}n!\delta_{mn}
\end{equation}
and the property that $H_n' = nH_{n-1}$ (Apell sequence), one can obtain closed form features for $d=1,2$ as shown in Table~\ref{table:fourier_features}. It is also known that the family of polynomial basis functions which are orthogonal and whose derivatives are orthogonal belong to one of three families, Jacobi, Laguerre or Hermite~\cite{webster1935orthogonal}. The extended support of the weight function of the Hermite basis, makes them well suited for additive modeling.

Although both these basis are complete, for practical purposes one has to use the first few basis. The quality of approximation depends on how well the underlying function can be approximated by these chosen basis,  for example, low degree polynomials are better represented by Hermite basis.  

\begin{table}
\begin{center}
\begin{tabular}{l|l}
Fourier & Hermite \\
$x \in [-1,1]$, $w(x) = 1$ & $x \in N(0,1)$, $w(x) = e^{-x^2/2}$\\
\hline \\
$\phi_n^1(x) = \{ \frac{\cos(n\pi x)}{n}, \frac{\sin(n\pi x)}{n} \}$ & $\phi_n^1(x) = \frac{H_n(x)}{\sqrt{nn!}} $\\\\
$\phi_n^2(x) = \{ \frac{\cos(n\pi x)}{n^2}, \frac{\sin(n\pi x)}{n^2} \}$ & $\phi_1^2(x) = \phi_1^1(x)$, $\phi_n^2(x) = \frac{H_n(x)}{\sqrt{n(n-1)n!}}$, $n > 1$ \\
\end{tabular}
\caption{\label{table:fourier_features} Fourier and Hermite encodings $\boldsymbol \phi^d$ for spline regression penalizing the $d$th derivative.}
\end{center}
\end{table}

\section{Additive Kernel Reproducing Kernel Hilbert Space \& Spline Embeddings}
\label{sec:add_rkhs}
\begin{figure*}
\begin{tabular}{c@{\hspace{0.05in}}c@{\hspace{0.0in}}c@{\hspace{0.05in}}c@{\hspace{0.0in}}c@{\hspace{0.05in}}c@{\hspace{0.0in}}c}
\includegraphics[width=0.135\linewidth]{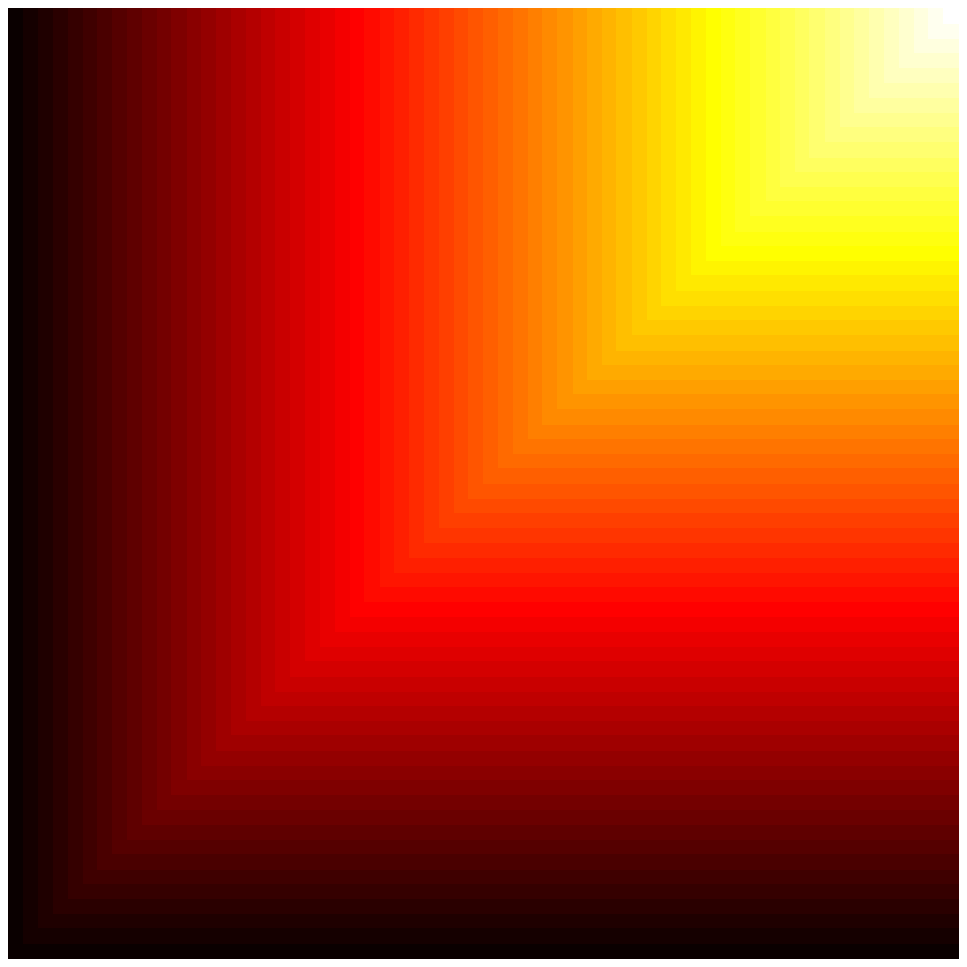} & 
\includegraphics[width=0.135\linewidth]{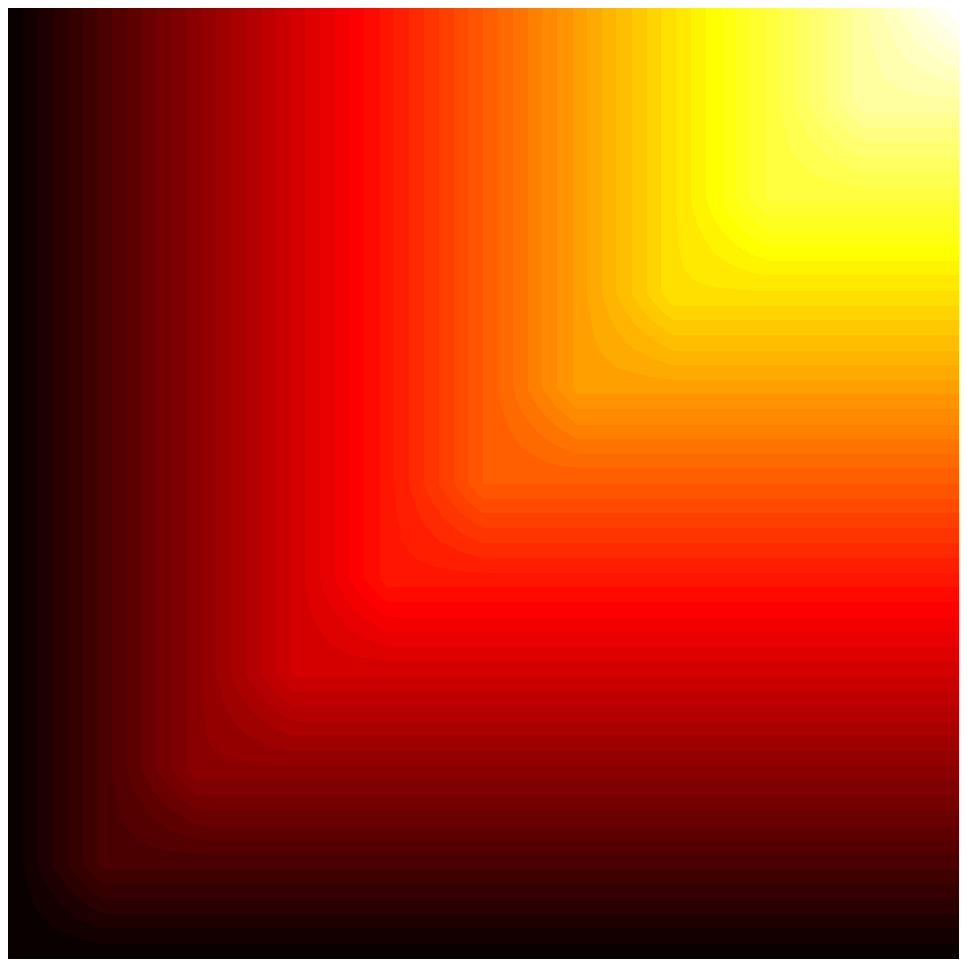} & 
\includegraphics[width=0.135\linewidth]{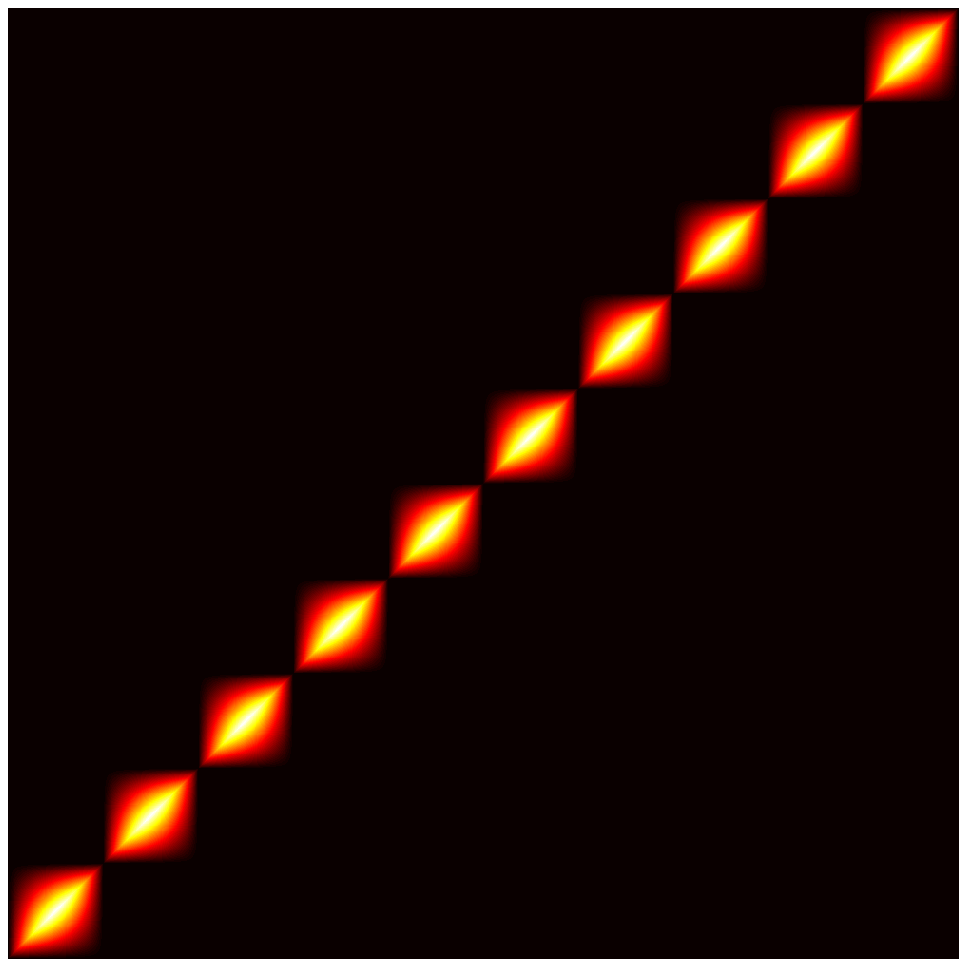} & 
\includegraphics[width=0.135\linewidth]{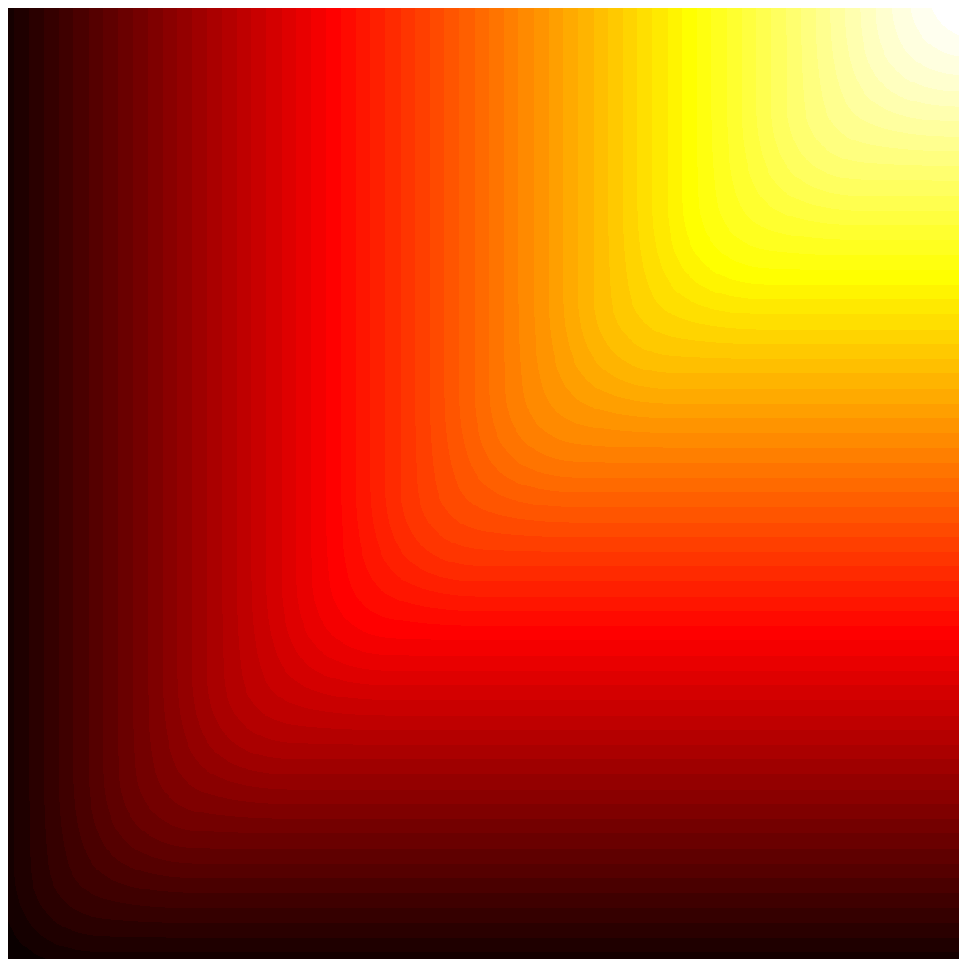} & 
\includegraphics[width=0.135\linewidth]{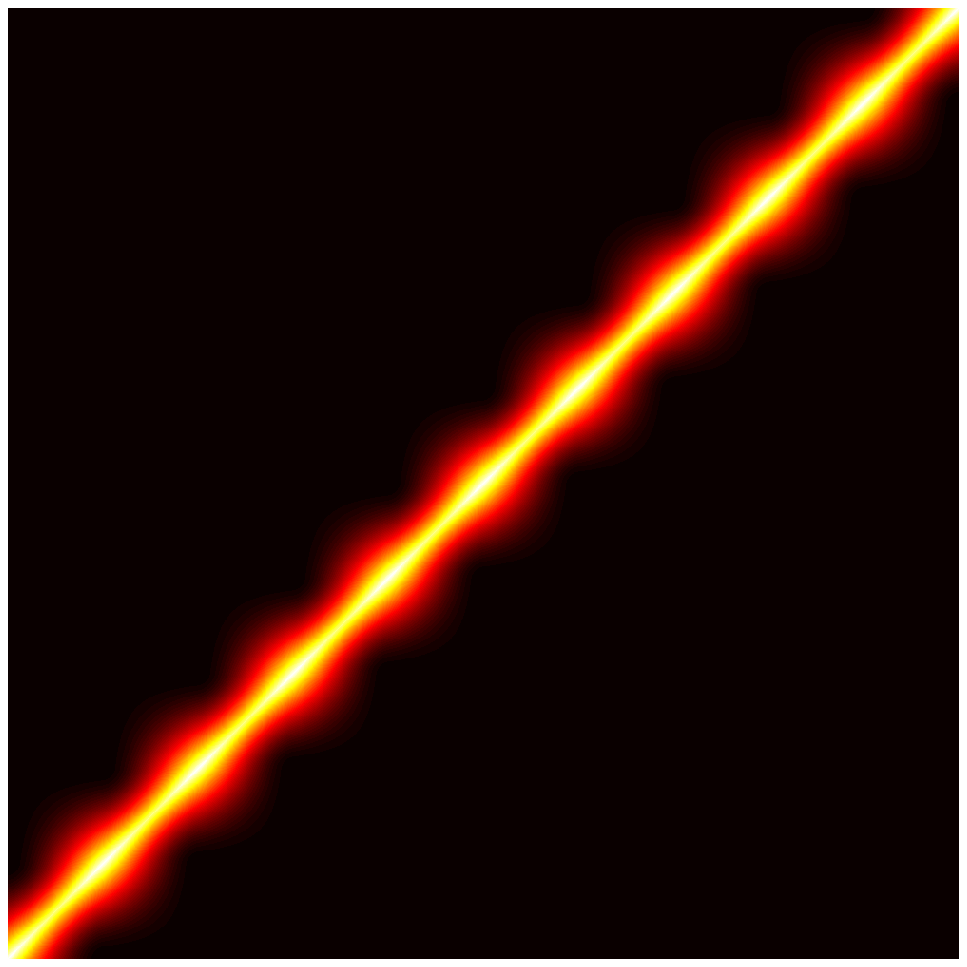} & 
\includegraphics[width=0.135\linewidth]{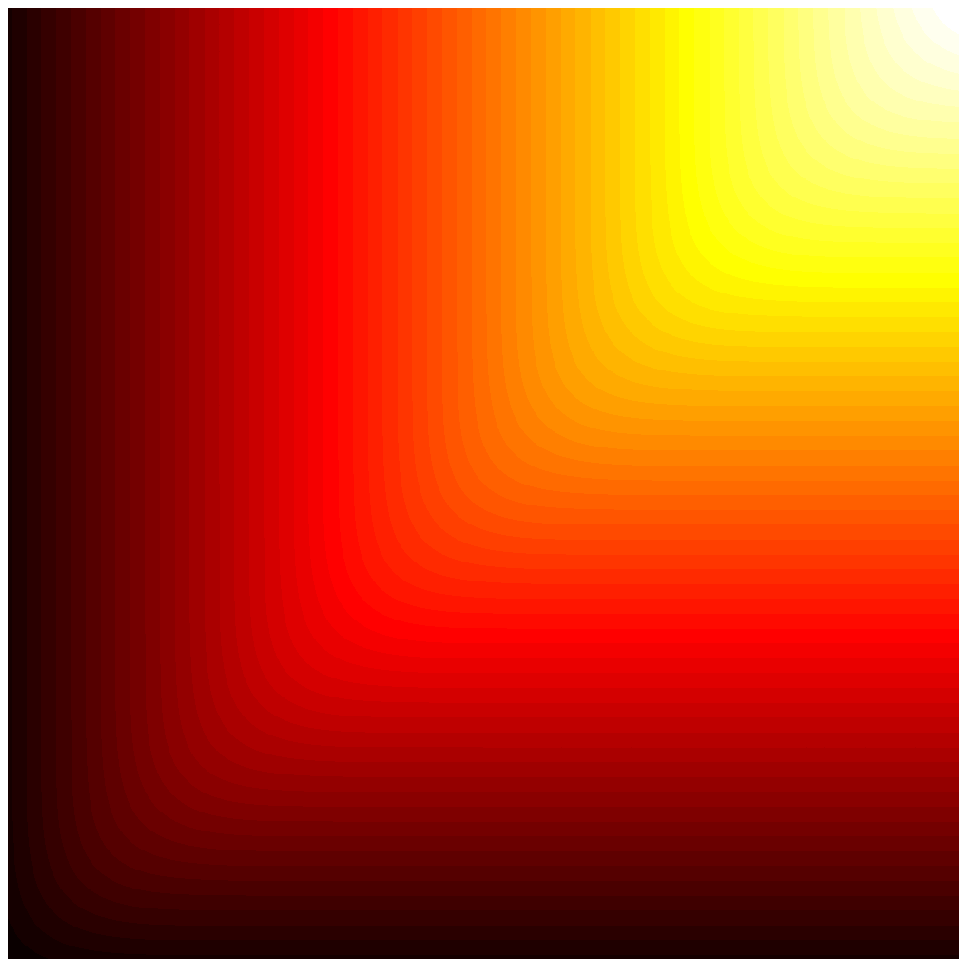} & 
\includegraphics[width=0.135\linewidth]{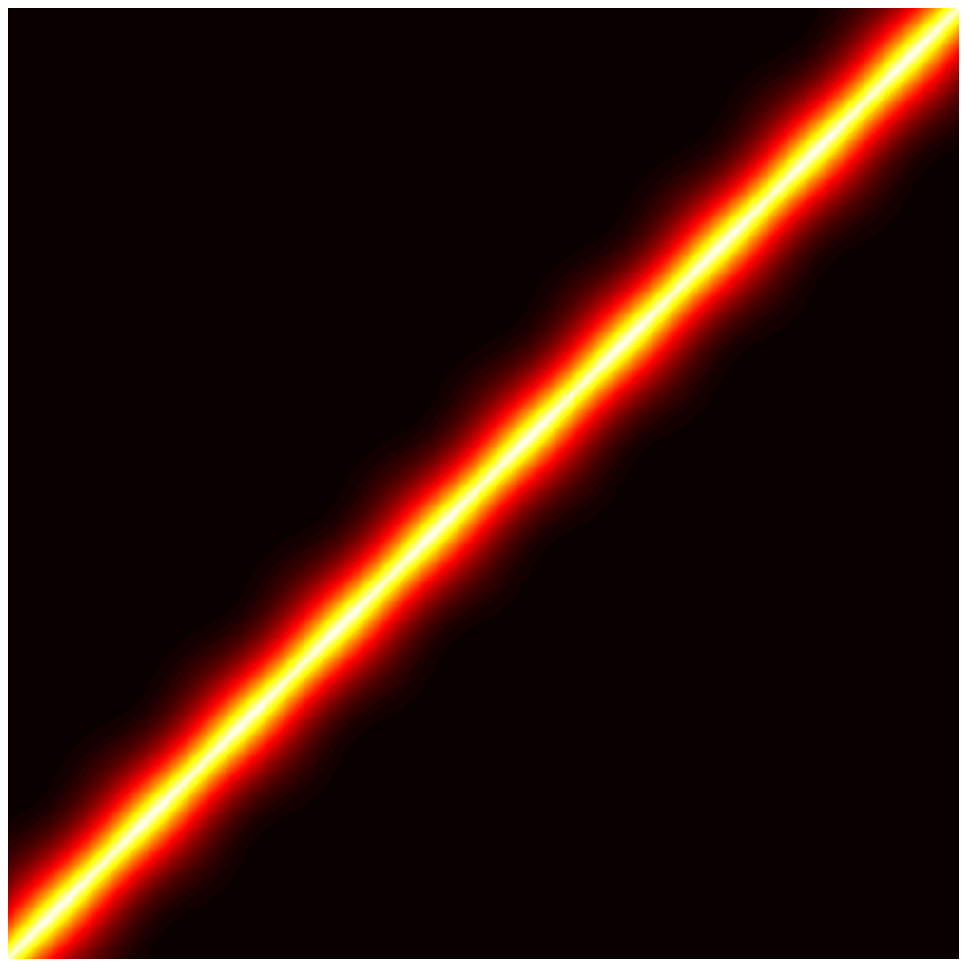} \\
$K_{\min}$ & $K_{1}^1$ & $K_{\min}-K_{1}^1$& $K_{2}^1$ & $K_{\min}-K_{2}^1$ & $K_{3}^1$ & $K_{\min}-K_{3}^1$\\ 
\end{tabular}
\caption{ \label{fig:approx_min_kernel}\textbf{Spline Kernels}. $K_{\min}(x,y), x,y \in [0,1]$ along with $K_r^1$ for $r=1,2,3$ corresponding to linear, quadratic and cubic B-Spline basis. Using uniformly spaced basis separated by $0.1$, these kernels closely approximate the $\min$ kernel as seen in the difference image. The approximation is exact when $|x-y| > 0.1r$. }
\end{figure*}

We begin by showing the close resemblance of the spline embeddings to the $\min$ kernel. To see this, let the features in $[0,1)$ be represented with $N+1$ uniformly spaced linear spline basis centered at $0, \frac{1}{N}, \frac{2}{N},\ldots, 1$. Let $r = \lfloor Nx \rfloor$ and let $\alpha = Nx -r$ . Then the features $\boldsymbol \phi(x)$ is given by $\phi_r(x) = 1-\alpha, \phi_{r+1}(x) = \alpha$ and the features $\boldsymbol \phi^1(x)$ for $D_1$ matrix is given by $\phi^1_i(x) = 1$, if $i \leq r$ and $\phi^1_r(x) = \alpha$. It can be seen that these features closely approximates the $\min$ kernel, i.e. 
\begin{equation}
\frac{1}{N}\boldsymbol \phi^1(x)'\boldsymbol \phi^1(y) \approx \min(x,y) + 1
\end{equation} 
The features $\boldsymbol  \phi^1(x) = D_1^{'-1} \boldsymbol  \phi(x)$ constructs a unary like representation where the number of ones equals the position of the bin of $x$. One can verify that for a B-spline basis of degree $r$ ($r=1,2,3$), the following holds:
\begin{equation}
\frac{1}{N}\boldsymbol \phi^1(x)' \boldsymbol \phi^1(y) = \min(x,y) + \frac{r+1}{2}, \mbox{if} |x-y| \geq \frac{r}{N}
\end{equation}
Define $K^r_{d}$ the kernel corresponding to a B-Spline basis of degree $r$ and regularization matrix $D_d$ as follows: 
\begin{equation}
	K^r_{d}(x,y) = \frac{1}{N}\boldsymbol \phi^d(x)' \boldsymbol \phi^d(y) - \frac{r+1}{2} = \frac{1}{N}\boldsymbol \phi(x)' D_d^{-1}D_d^{'-1} \boldsymbol \phi(y) - \frac{r+1}{2}
\end{equation}
Figure~\ref{fig:approx_min_kernel} shows $K^1_r$ for $r=1,2,3$ corresponding to a linear, quadratic and cubic B-Spline basis. In a recent paper, Maji and Berg~\cite{maji2009max}, propose to use linear spline basis and a $D_1$ regularization, to train approximate intersection kernel SVMs, which in turn approximate arbitrary additive classifiers. Our features can be seen as a generalization to this work which allows arbitrary spline basis and regularizations. 

B-Splines are closely related to the truncated polynomial kernel~\cite{wahba1990spline,pearce2006penalized} which consists of uniformly spaced knots $\tau_1, \ldots, \tau_n$ and truncated polynomial features:
\begin{equation}
\phi_i(x) = (x-\tau_i)^p_+
\end{equation}
However these features are not as numerically stable as B-Spline basis (see~\cite{eilers2005splines} for an experimental comparison). Truncated polynomials of degree $k$ corresponds to a B-Spline basis of degree $k$ and $D_{k+1}$ regularization, i.e, same as $K_k^{k+1}$ kernel, when the knots are uniformly spaced. This is because B-Splines are derived from truncated polynomial basis by repeated application of the difference matrix $D_1$\cite{de2001practical}. As noted by the authors in~\cite{eilers2005splines}, one of the advantages of the P-Spline formulation is that is decouples the order of regularization and B Spline basis. Typically $D_1$ regularization provides sufficient smoothness in our experiments.

\section{Optimizations for Efficient Learning for B-Spline embeddings}
\label{sec:sparse_embeddings}
For B-spline basis one can exploit the sparsity to speed-up linear solvers. The classification function is based on evaluating $\mathbf{w}'D_d^{'-1}\phi(x)$. Most methods for training linear methods are based on evaluating the classifier and updating the classifier if the classification is incorrect. Since the number of such evaluations are larger than the number of updates, it is is much more efficient to maintain $\mathbf{w}_d=D_d^{-1}\mathbf{w}$ and use sparse vector multiplication. Updates to the weight vector $\mathbf{w}$ and $\mathbf{w}_d$ for various gradient descent algorithms look like:  
\begin{equation}
\mathbf{w}^t \leftarrow \mathbf{w}^{t-1} - \eta D_d^{'-1}\phi(x^k),~~~~\mathbf{w}_d^t = \mathbf{w}_d^{t-1} - \eta L_d \phi(x^k) 
\end{equation}
Where $\eta$ is a step and $L_d = D_d^{-1}D_d^{'-1}$. Unlike the matrix $D_d'D_d$, the matrix $L_d$ is a dense, and hence the updates to $\mathbf{w}_d$ may change all the entries of $\mathbf{w}_d$. However, one can compute $L_d\boldsymbol \phi(x)$ in $2dn$ steps instead of $n^2$ steps, exploiting the simple form of $D_1^{'-1}$. Initialize $a_i = \phi_i(x)$ then repeat step A, $d$ times followed by step B, $d$ times to compute $L_d\phi(x)$.
\begin{eqnarray*}
	\mbox{step A} &:& a_i  = a_i + a_{i+1}, i = n-1 \mbox{ to } 1\\
	\mbox{step B} &:& a_i   = a_i + a_{i-1}, i = 2 \mbox{ to } n 
\end{eqnarray*}

%\subsection{Encoding sparse features}
%For input features that are sparse and non-negative, which often arise in "bag-of-words" representations of text documents or images, it is important to preserve the sparsity in the embeddings for computational and memory efficiency. Formally, we need the property that $\boldsymbol \phi(0) = \mathbf{0}$. 

%For the B-Spline basis one can achieve this by placing the leftmost basis function away from $0$, such that $x=0$ has no support on any of the basis functions. For the generalized Fourier based features one could consider an expansion using just the odd functions $\left( f(-x) = -f(x) \right)$. For Fourier features this corresponds to using just the $\sin(n\pi x)$ basis, or the odd Hermite polynomials, i.e, $H_1, H_3,\ldots,H_{2n+1}$.
\section{Experiments}
Often on large datasets consisting of very high dimensional features, to avoid the memory bottleneck, one may compute the encodings in the inner loop of the training algorithm. We refer this to as the ``online" method. Our solver is based on \texttt{LIBLINEAR}, but can be easily used with any other solver. The custom solver allows us to exploit the sparsity of embeddings (Section~\ref{sec:sparse_embeddings}). A practical regularization is $D_0 = I$ with the B-Spline embeddings, where $I$ is the identity matrix, which leads to sparse features. This makes it difficult to estimate the weights on the basis functions which have no data, but one can use a higher order B-Spline basis, to somewhat mitigate this problem. 

We present image classification experiments on two image datasets, \texttt{MNIST}~\cite{lecun1998mnist} and Daimler Chrysler (\texttt{DC}) pedestrians~\cite{munder2006experimental}. On these datasets SVM classifiers based on histogram intersection kernel outperforms a linear SVM classifier~\cite{maji2009max,maji09fast}, when used with features based on a spatial pyramid of histogram of oriented gradients~\cite{dalal2005histograms, lazebnik2006beyond}. We obtain the features from the author's website for our experiments. The \texttt{MNIST} dataset has $60,000$ instances and the features are $2172$ dimensional and dense, leading to $130,320,000$ non-zero entries. The \texttt{DC} dataset has three training sets and two test sets. Each training set has $19,800$ instances  and the features are $656$ dimensional and dense, leading to $12,988,800$ entries. These sizes are typical of image datasets and training kernel SVM classifiers often take several hours on a single machine. 

\begin{figure*}
\begin{tabular}{c@{\hspace{0.0in}}c@{\hspace{0.0in}}c@{\hspace{0.0in}}c}
\includegraphics[height=1.2in]{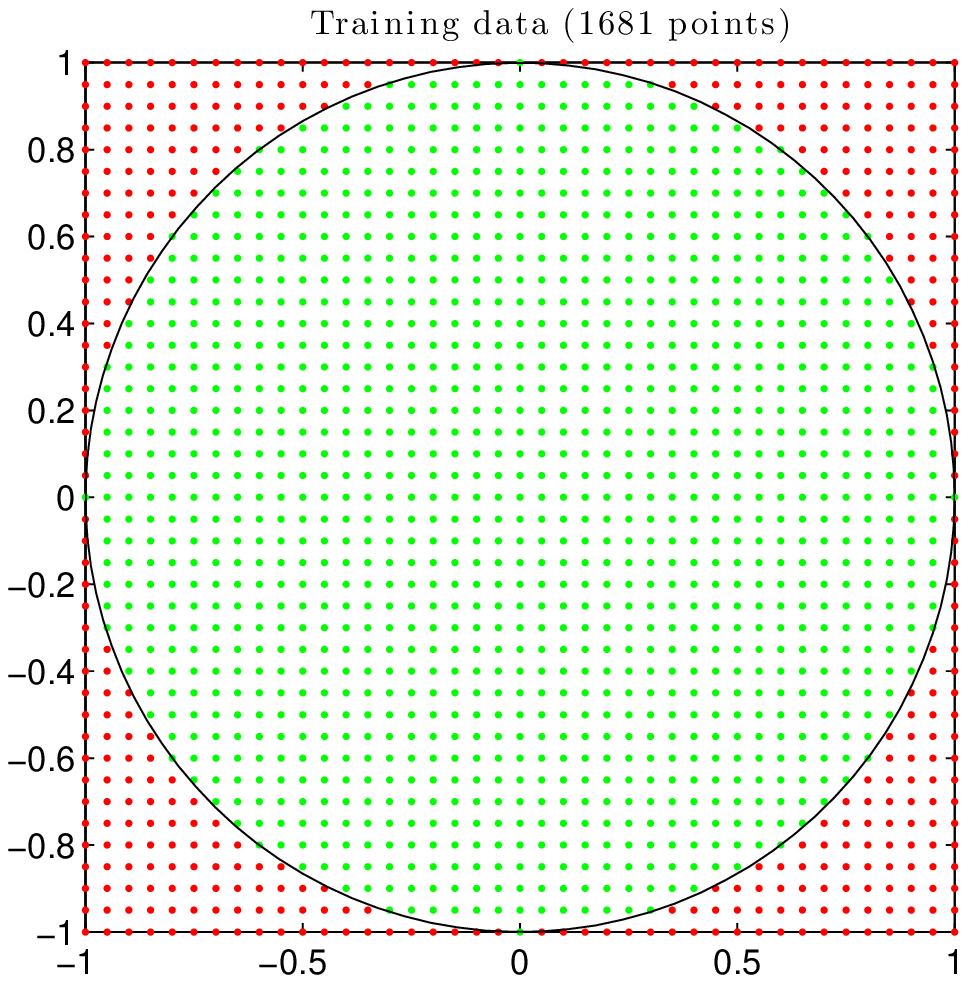} & 
\includegraphics[height=1.2in]{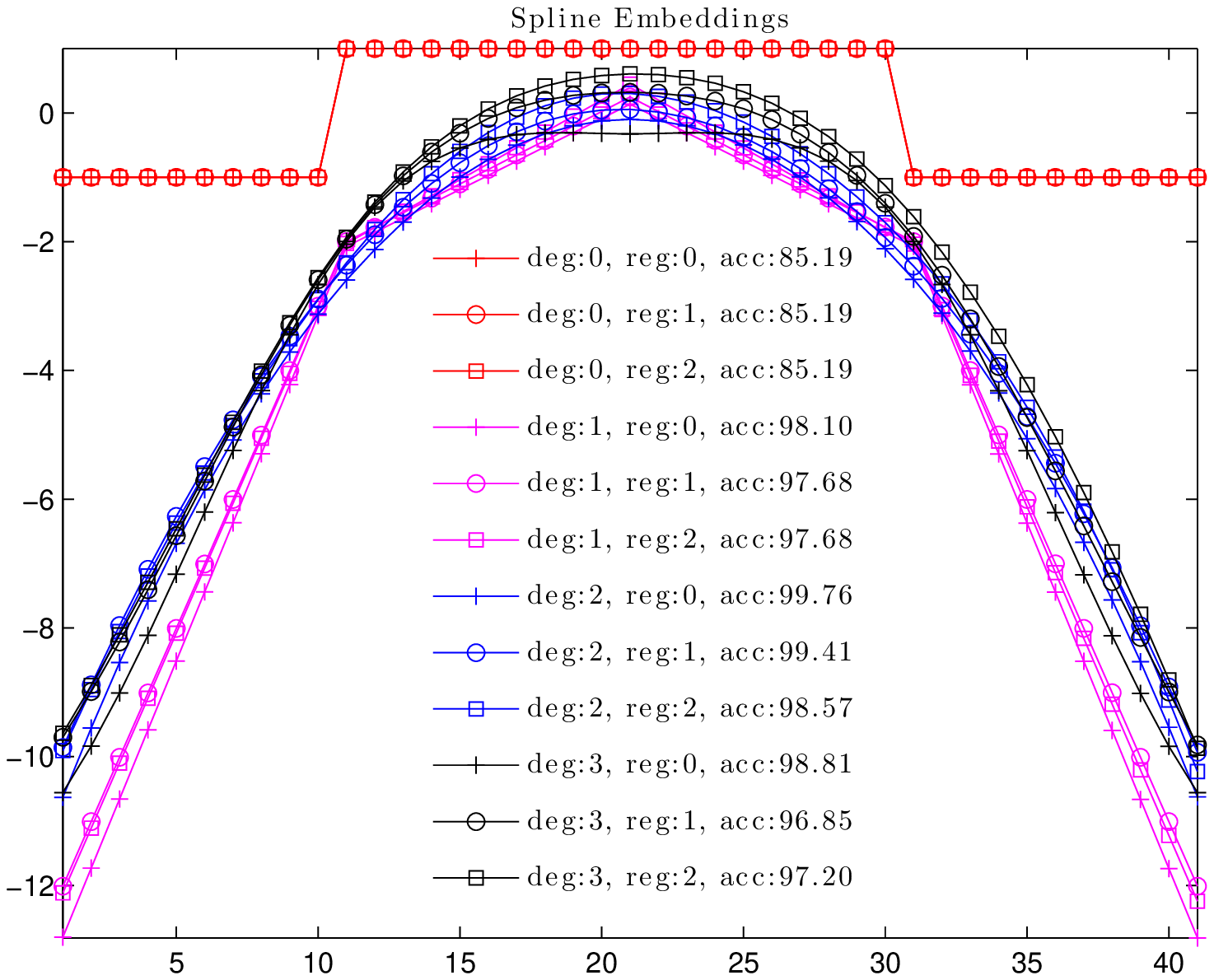} &
\includegraphics[height=1.2in]{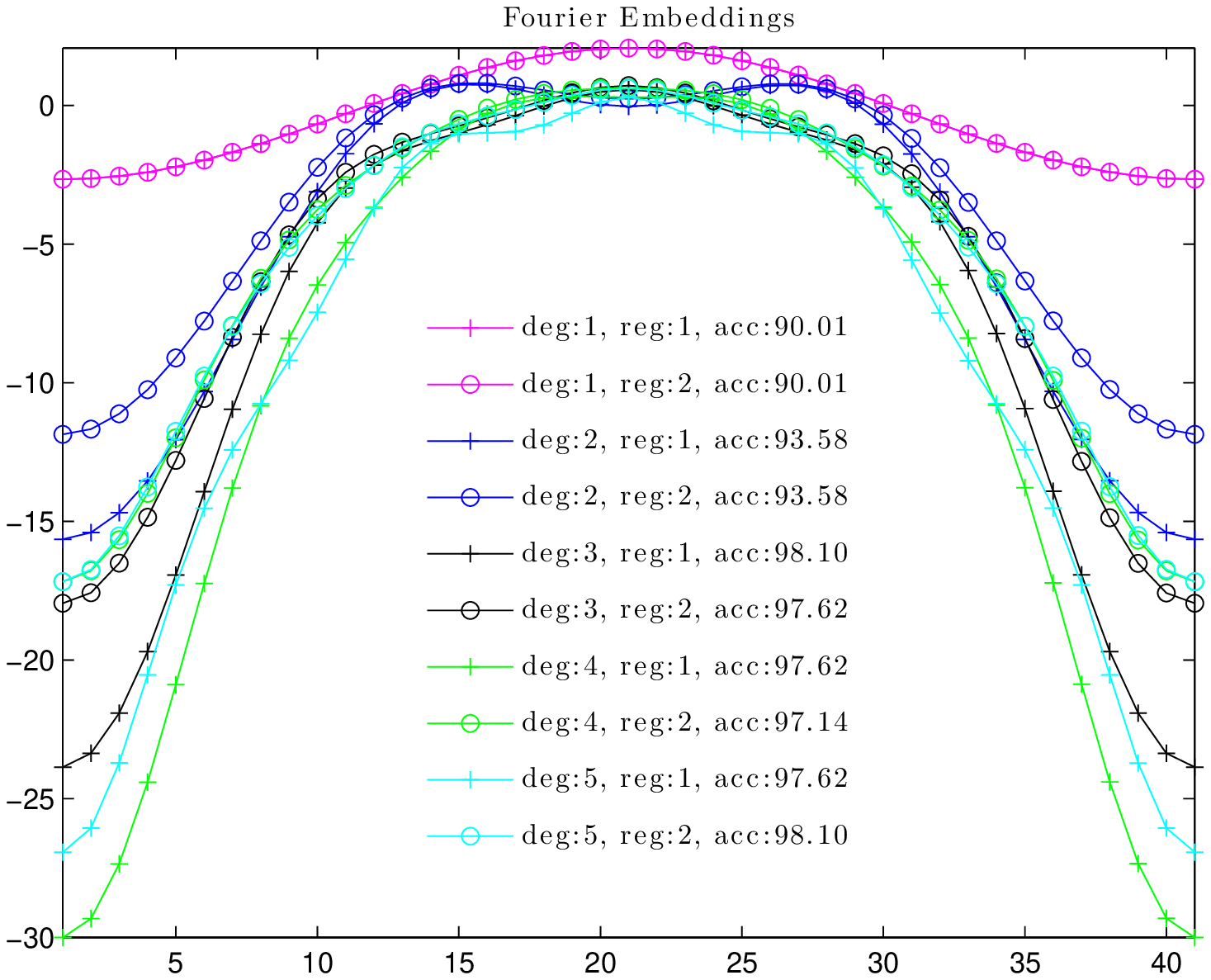} & 
\includegraphics[height=1.2in]{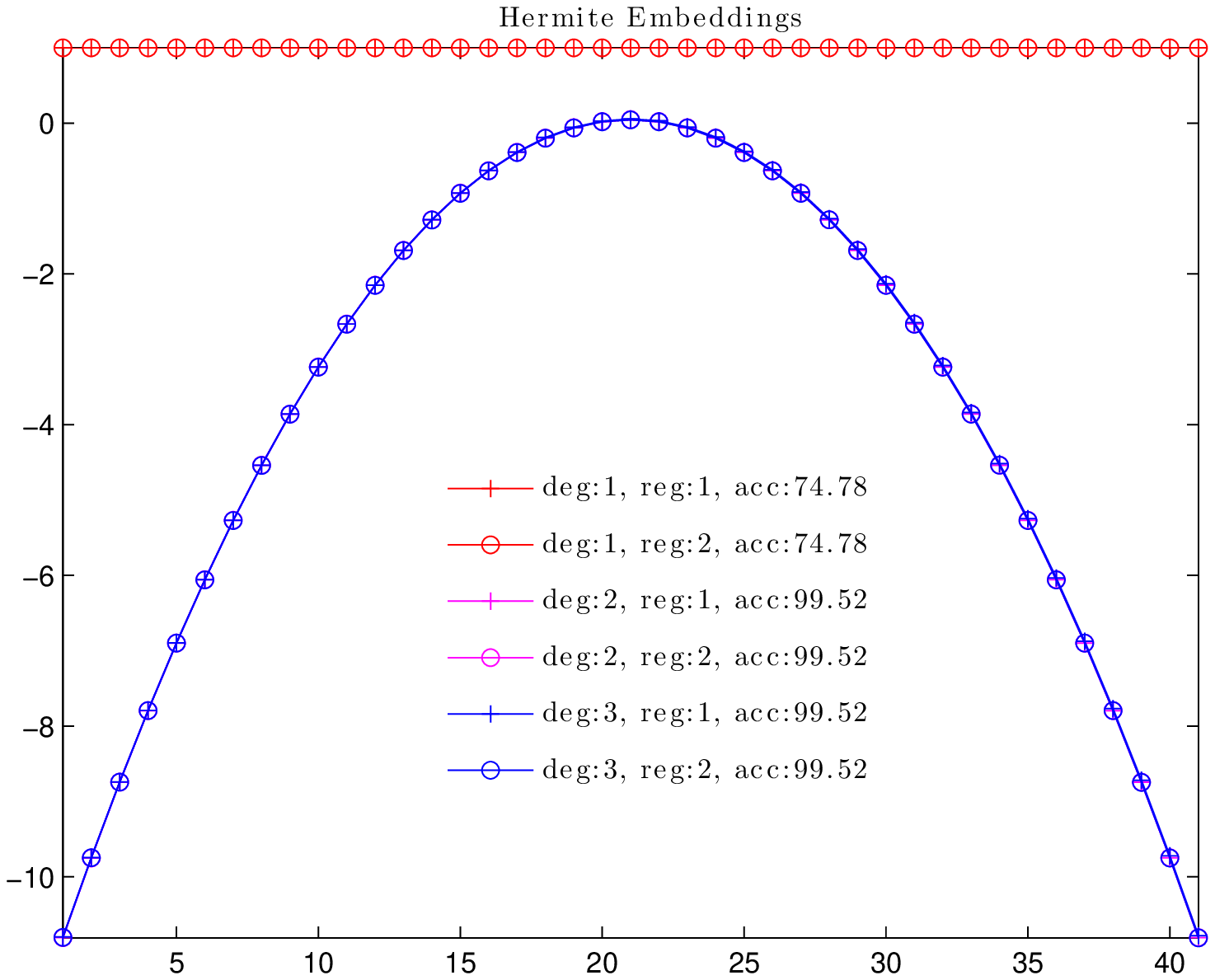}\\ 
(a) & (b) & (c) & (d) \\
\end{tabular}
\caption{\label{fig:toy-dataset}(a) 2D data generated according to $x^2 + y^2 \leq 1$. (b) B-Spline fits along $x$ for various degrees  and regularizations and $4$ uniformly spaced bins in $[-1,1]$. (c) Fourier fits along $x$ using $\sin,\cos$ features for various frequencies and regularizations. (d) Hermite fits along $x$ for various degrees and regularizations. }
\vspace{-0.1in}
\end{figure*}

\paragraph{Toy Dataset.} 
The points are sampled uniformly on a 2D grid $[-1,1]\times[-1,1]$ with the points satisfying $x^2 + y^2 \leq 1$ in the positive class and others as negative. Figure~\ref{fig:toy-dataset}(b), shows the fits on the data along $x$ (or $y$) dimension using $4$ uniformly spaced B-spline basis of various degrees and regularizations. The quadratic and cubic splines offer smoother fits of the data. Figure~\ref{fig:toy-dataset}(c,d) shows the learned functions using Fourier and Hermite embeddings of various degrees respectively. 

\paragraph{Effect of B-Spline parameter choices.}Table~\ref{tab:DCped}, shows the accuracy and training times as a function of the number of bins, regularization degree ($D_r, r=0,1,2$) and the B-Spline basis degree ($d=1,2,3)$ on the first split of the DC pedestrian dataset. We set $C=1$ and the bias term $B=1$ for training all the models. On this dataset we find that $r=0,1$ is more accurate than $r=3$ and is significantly faster. \emph{In further experiments we only include the results of $r=0,1$ and $d=1,3$.} In addition, setting the regularization to zero ($r=0$), leads to very sparse features and can be used directly with any linear solver which can exploit this sparsity. The training time for B-Splines scales sub-linearly with the number of bins, hence better fits of the functions can be obtained without much loss in efficiency.

\begin{table*}
\begin{center}
\begin{tabular}{|c|ccc|}
\hline
            &\multicolumn{3}{c|}{Regularization} \\
Degree & $D_0$ & $D_1$ & $D_2$ \\
\hline
            &\multicolumn{3}{c|}{$\mathbf{5}$ \textbf{bins}} \\
1 & $\mathbf{06.60s}$ $(89.55\%)$ & $\mathbf{20.27s}$ $(89.68\%)$ & $\mathbf{041.60}s$ $(89.93\%)$ \\
2 & $08.74s$ $(\mathbf{90.45\%})$ & $30.47s$ $(\mathbf{90.20}\%)$ & $080.25s$ $(\mathbf{89.94}\%)$\\
3 & $11.68s$ $(90.03\%)$ & $49.85s$ $(89.93\%)$ & $143.50s$ $(88.57\%)$ \\
\hline
            &\multicolumn{3}{c|}{$\mathbf{10}$ \textbf{bins}} \\
1 & $\mathbf{05.61s}$ $(90.42\%)$ & $\mathbf{23.06}s$ $(90.86\%)$ & $\mathbf{077.99}s$ $(\mathbf{89.43}\%)$ \\
2 & $08.10s$ $(\mathbf{90.69}\%)$ & $29.97s$ $(\mathbf{90.73\%})$ & $126.03s$ $(89.23\%)$\\
3 & $11.59s$ $(90.48\%)$ & $42.26s$ $(90.67\%)$ & $193.47s$ $(89.14\%)$ \\
\hline
            &\multicolumn{3}{c|}{$\mathbf{20}$ \textbf{bins}} \\
1 & $\mathbf{05.96s}$ $(90.23\%)$ & $\mathbf{32.43}s$ $(\mathbf{91.20\%})$ & $\mathbf{246.87}s$ $(\mathbf{89.06}\%)$ \\
2 & $07.26s$ $(90.34\%)$ & $34.99s$ $(91.10\%)$ & $328.32s$ $(88.89\%)$\\
3 & $10.08s$ $(\mathbf{90.39\%})$ & $42.88s$ $(91.00\%)$ & $429.57s$ $(88.92\%)$ \\
\hline
\end{tabular}
\caption{\label{tab:DCped} \small{\textbf{B-Spline parameter choices on DC dataset.} Training times and accuracies for various B-Spline parameters on the first split of DC dataset. In comparison training a linear model using \texttt{LIBLINEAR} takes $3.8s$ and achieves $81.5\%$ accuracy. With $D_0$ regularization, higher degree splines are more accurate, but with $D_1$ regularization linear spline basis are as accurate as quadratic/cubic when the number of bins are large. Training using $D_0$ regularization is about than $4\times$ faster than $D_1$. Higher order regularization seems unnecessary. \emph{Note that this implementation computes the encodings on the fly hence has the same memory overhead as} \texttt{LIBLINEAR}.}}
\end{center}
\vspace{-0.2in}
\end{table*}
\paragraph{Effect Fourier embedding parameter choices.} Table~\ref{tab:dcped_fourier}, shows the accuracy and training times for various Fourier embeddings on \texttt{DC} dataset. Before computing the generalized Fourier features, we first normalize the data in each dimension to $[-1,1]$ using:
\begin{equation}
	x \leftarrow \frac{x -\mu}{\delta}, \mbox{where, } \mu = \frac{x_{\max} + x_{\min}}{2},  \delta = \frac{x_{\max} - x_{\min}}{2}
\end{equation}
We precompute the features and use \texttt{LIBLINEAR} to train various models, since it is relatively more expensive to compute the features online. In this case the training times are similar to that of B-Spline models. However, precomputing and storing the features may not be possible on very large scale datasets. 

\begin{table*}
\begin{tabular}{|c|cc|cc|cc|cc|}
\hline
             &\multicolumn{4}{|c|}{Fourier} &\multicolumn{4}{|c|}{Hermite}\\
             \cline{2-9}
             &\multicolumn{2}{|c|}{$r=1$}  &\multicolumn{2}{|c|}{$r=2$}  &\multicolumn{2}{|c|}{$r=1$}  &\multicolumn{2}{|c|}{$r=2$} \\
Degree & Accuracy & Time & Accuracy & Time & Accuracy & Time & Accuracy & Time\\
\hline
1 	& $88.94\%$	&$\mathbf{07.0s}$ & $88.94\%$	&$\mathbf{07.0s}$ & $84.17\%$	&$\mathbf{02.8s}$ & $84.17\%$	&$\mathbf{02.8s}$ \\
2 	& $89.59\%$&$10.2s$ & $89.64\%$	&$10.2s$ & $88.01\%$	&$04.6s$ & $88.01\%$	&$04.6s$ \\
3 	& $88.99\%$&$12.7s$ & $89.77\%$	&$12.8s$ & $88.22\%$	&$07.7s$ & $88.70\%$	&$09.9s$ \\
4 	& $\mathbf{89.77}\%$	&$16.0s$ & $\mathbf{89.84}\%$	&$15.9s$ & $\mathbf{89.00\%}$	&$12.6s$ & $\mathbf{89.05\%}$	&$11.9s$ \\
\hline
\end{tabular}
\caption{\label{tab:dcped_fourier}{\small \textbf{Fourier embeddings on DC dataset}. Accuracies and training times (encoding + training) for various Fourier embeddings on the first split of the DC pedestrian dataset. \emph{Note that the methods are trained by first encoding the features and training a linear model using} \texttt{LIBLINEAR}.}}
\vspace{-0.2in}
\end{table*}

\vspace{-0.1in}
\paragraph{Comparison of various additive models.} { Table~\ref{tab:dcped_full}, shows the accuracy and training times of various additive models compared to linear and the more expensive $\min$ kernel SVM on all the $6$ training and test set combinations of \texttt{DC} dataset. The optimal parameters were found on the first training and test set. The additive models are up to $50\times$ faster to train and are as accurate as the $\min$ kernel SVM. The B-Spline additive models significantly outperform a linear SVM on this dataset at the expense of small additional training time. 

Table~\ref{tab:mnist_full} shows the accuracies and training times of various additive models on the \texttt{MNIST} dataset. We train one-vs-all classifiers for each digit and the classification scores are normalized by passing them through a logistic. During testing, each example given the label of the classifier with the highest response. The optimal parameters for training were found using 2-fold cross validation on the training set. Once again the additive models significantly outperform the linear classifier and closely matches the accuracy of $\min$ kernel SVM, while being $50\times$ faster. }

\vspace{-0.1in}
\begin{table*}
\begin{center}
\begin{tabular}{|r|c|r|c|}
\hline
Method 					& Test Accuracy 			& \multicolumn{2}{|c|}{Training Time} \\
\hline
SVM (linear) + \texttt{LIBLINEAR} 	& $81.49$ ($1.29$) 			& \multicolumn{2}{|c|}{$3.8$s}\\
SVM ($\min$)  + \texttt{LIBSVM} 		& $89.05$ ($1.42$)			& \multicolumn{2}{|c|}{$363.1$s}\\		
\hline
						&        					& online & batch \\
\hline
B-Spline $(r=0,d=1,n=05)$  		& $88.51$ ($1.35$)			& $\mathbf{5.9}$s      & - \\
B-Spline $(r=0,d=3,n=05)$  		& $89.00$ ($1.44$)			& $10.8$s   & - \\
B-Spline $(r=1,d=1,n=10)$ 		& $\mathbf{89.56}$ ($1.35$)	& $17.2$s    & - \\
B-Spline $(r=1,d=3,n=10)$ 		& $89.25$ ($1.39$)			& $19.2$s    & - \\
Fourier  $(r=1,d=2)$ 		& $88.44$ ($1.43$)			& $159.9$s  & $12.7$s ($4\times$ memory)\\
Hermite $(r=1,d=4)$ 		& $87.67$ ($1.26$)			& $35.5$s    & $12.6$s ($4\times$ memory)\\
\hline
\end{tabular}
\vspace{-0.1in}
\caption{\label{tab:dcped_full} {\small \textbf{Training times and test accuracies of various additive classifiers on DC dataset}. The online training method compute the encodings on the fly and the batch method computes the encoding once and uses \texttt{LIBLINEAR} to train a linear classifier. For B-Splines the online method is faster hence we omit the training times for the batch method. All the additive classifiers outperform the linear SVM, while being up to $50\times$ faster than $\min$ SVM. }}
\end{center}
\vspace{-0.3in}
\end{table*}\vspace{-0.1in}
\begin{table*}
\begin{center}
\begin{tabular}{|r|c|r|}
\hline
Method 						& Test Error 			& Training Time \\
\hline
SVM (linear) + \texttt{LIBLINEAR} 	& $1.44\%$  			& $6.2$s\\
SVM ($\min$)  + \texttt{LIBSVM} 	& $0.79\%$ 			& $\sim2.5$ hours\\		
\hline
%B-Spline $(r=0,d=1,n=05)$  		& $0.93\%$			& $84.9$s      \\
%B-Spline $(r=0,d=1,n=10)$  		& $0.96\%$			& $47.6$s      \\
B-Spline $(r=0,d=1,n=20)$  		& $0.88\%$ 			& $31.6$s   \\
%B-Spline $(r=0,d=1,n=40)$  		& $1.06\%$ 			& $32.7$s   \\
%B-Spline $(r=0,d=3,n=05)$  		& $-\%$			         & $84.9$s      \\
%B-Spline $(r=0,d=3,n=10)$  		& $1.04\%$			& $87.7$s      \\
B-Spline $(r=0,d=3,n=20)$  		& $0.86\%$ 			& $51.6$s   \\
%B-Spline $(r=1,d=1,n=20)$  		& $0.87\%$ 			& $127.4$s   \\
B-Spline $(r=1,d=1,n=40)$  		& $\mathbf{0.81\%}$ 			& $157.7$s   \\
%B-Spline $(r=1,d=1,n=100)$  		& $0.81\%$ 			& $286.7$s   \\
%B-Spline $(r=1,d=3,n=20)$  		& $0.85\%$ 			& $261.4$s   \\
B-Spline $(r=1,d=3,n=40)$  		& $0.82\%$ 			& $244.9$s   \\
Hermite $(r=1,d=4)$ 			& $1.06\%$		& $358.6$s   \\
\hline
\end{tabular}
\caption{\label{tab:mnist_full} {\small \textbf{Test error and mean training times per digit for various additive classifiers on MNIST}. For B-Splines the online method is faster hence we omit the training times for the batch method. All the additive classifiers outperform the linear SVM, while being up to $50\times$ faster than $\min$ SVM. }}
\end{center}
\end{table*}
\section{Conclusion}
\vspace{-0.1in}
{\small We have proposed a family of embeddings which enable efficient learning of additive classifiers. We advocate the use B-Splines based embeddings because they are are efficient to compute and are sparse, which enables us to train these models with a small memory overhead by computing the embeddings on the fly even when the number of basis are large and can be seen as a generalization of~\cite{maji2009max}. Generalized Fourier features are low dimensional, but are expensive to compute and so are more suitable if the projected features can be precomputed and stored. The proposed classifiers outperform linear classifiers and match the significantly more expensive kernel SVM classifiers at a fraction of the training time. On both the \texttt{MNIST} and \texttt{DC} datasets, linear B-Spline and $D_1$ regularization works the best and closely approximates the learning problem of $\min$ kernel SVM. Higher degree splines are useful when used with $D_0$ regularization, have even faster training times but worse accuracies than $D_1$ regularization. The code for training various spline models proposed in the paper has been packaged as a library, \texttt{LIBSPLINE}, which will be released upon the publication of this paper.}   
\bibliographystyle{ieee}
\bibliography{embeddings}

\end{document}